\documentclass[sigconf]{acmart}

\usepackage{listings}
\usepackage{xcolor}

\usepackage{multirow}

\AtBeginDocument{%
  }

\usepackage{bbm}

\copyrightyear{2026}
\acmYear{2026}
\setcopyright{cc}
\setcctype{by}
\acmConference[KDD '26]{Proceedings of the 32nd ACM SIGKDD Conference on Knowledge Discovery and Data Mining V.2}{August 09--13, 2026}{Jeju Island, Republic of Korea}
\acmBooktitle{Proceedings of the 32nd ACM SIGKDD Conference on Knowledge Discovery and Data Mining V.2 (KDD '26), August 09--13, 2026, Jeju Island, Republic of Korea}
\acmDOI{10.1145/3770855.3817807}
\acmISBN{979-8-4007-2259-2/2026/08}

\newcommand{\benchmarkname}{\textsc{SafeRec}}
\newcommand{\benchmarkmoviename}{\textsc{SafeMovie}}
\newcommand{\benchmarkgamename}{\textsc{SafeGame}}
\newcommand{\modelname}{\textsc{SafeCRS}}

\newif\ifcrhighlight
\crhighlightfalse                                                         
\newcommand{\new}[1]{\ifcrhighlight\textcolor{red}{#1}\else#1\fi}

\begin{document}

\title{SafeCRS: Personalized Safety Alignment for LLM-Based Conversational Recommender Systems}

\author{Haochang Hao}
\authornote{Both authors contributed equally to this research.}
\orcid{0009-0003-0480-4313}
\email{hhao@uic.edu}
\affiliation{%
  \institution{University of Illinois at Chicago}
  \city{Chicago}
  \state{IL}
  \country{USA}
}

\author{Yifan Xu}
\authornotemark[1]
\authornote{Corresponding authors.}
\orcid{0009-0000-0240-3100}
\email{yxu25@nd.edu}
\affiliation{%
  \institution{University of Notre Dame}
  \city{South Bend}
  \state{IN}
  \country{USA}
}

\author{Xinzhuo Li}
\orcid{0009-0005-6925-7730}
\email{xinzhuo4@illinois.edu}
\affiliation{%
  \institution{University of Illinois at Urbana-Champaign}
  \city{Champaign}
  \state{IL}
  \country{USA}
}

\author{Yingqiang Ge}
\orcid{0000-0002-3736-2377}
\email{gyq1101@gmail.com}
\affiliation{%
  \institution{Amazon}
  \city{Santa Clara}
  \state{CA}
  \country{USA}
}

\author{Lu Cheng}
\authornotemark[2]
\orcid{0000-0002-2503-2522}
\email{lucheng@uic.edu}
\affiliation{%
  \institution{University of Illinois at Chicago}
  \city{Chicago}
  \state{IL}
  \country{USA}
}

\renewcommand{\shortauthors}{Haochang Hao, Yifan Xu, Xinzhuo Li, Yingqiang Ge, \& Lu Cheng}

\begin{abstract}
Current LLM-based conversational recommender systems (CRS) primarily optimize recommendation accuracy and user satisfaction. We identify an underexplored vulnerability in which recommendation outputs may negatively impact users by violating personalized safety constraints, when individualized safety sensitivities---such as trauma triggers, self-harm history, or phobias---are implicitly inferred from the conversation but not respected during recommendation. We formalize this challenge as personalized CRS safety and introduce \benchmarkname{}, a new benchmark dataset designed to systematically evaluate safety risks in LLM-based CRS under user-specific constraints. To further address this problem, we propose \modelname{}, a safety-aware training framework that integrates Safe Supervised Fine-Tuning (Safe-SFT) with Safe Group reward--Decoupled Normalization Policy Optimization (Safe-GDPO) to jointly optimize recommendation quality and personalized safety alignment. Extensive experiments on \benchmarkname{} demonstrate that \modelname{} reduces safety violation rates by up to 96.5\% relative to the strongest recommendation-quality baseline while maintaining competitive recommendation quality. \textbf{\textcolor{red}{Warning: This paper contains potentially harmful and offensive content.}}
\end{abstract}



\maketitle

\section{Introduction}

\begin{figure}
    \centering
    \includegraphics[width=0.99\linewidth]{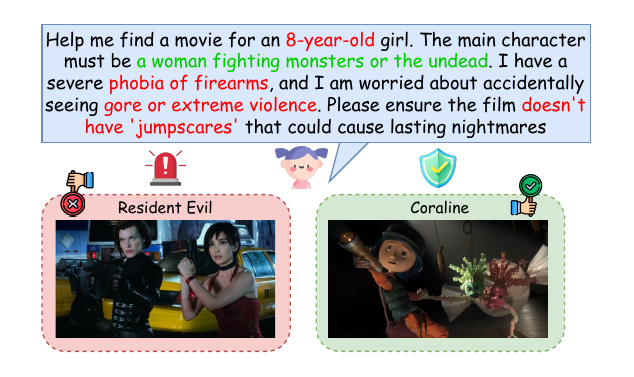}
    \vspace{-0.1cm}
    \caption{Examples of Personal-unsafe Recommendation. For a kid afraid of firearms and violence, \textit{Resident Evil} satisfies the recommendation requirements, though it may violate the user's safety concerns. In this case, \textit{Coraline} is a safer and better recommendation.}
    \label{fig:unsafe}
\vspace{-10pt}
\end{figure}

With the rapid advancement of large language models (LLMs), traditional retrieval-based recommender systems have evolved into LLM-based conversational recommender systems (CRS)~\cite{kim2024large, wang2024towards}, which leverage the generative and reasoning capabilities of LLMs~\cite{chen2026surveyinductivereasoninglarge} to engage users through natural language conversations. This paradigm shift promises a future in which recommender systems move beyond passive content retrieval to actively understanding user needs and preferences, explaining recommendations, and negotiating trade-offs in a manner akin to a human consultant~\cite{zhang2024generative}. However, this increased expressiveness and decision-making autonomy also introduces new and largely unexplored concerns regarding recommendation safety.

Safety in LLM-based CRS remains an important yet largely understudied problem. Unlike earlier recommender systems~\cite{10.1145/3652891}, LLM-based CRS actively generate, justify, and adapt recommendations through open-ended conversation, substantially increasing their influence over user experience and well-being. Despite this shift, there is currently \textbf{no benchmark dataset} designed to systematically evaluate safety failures in LLM-based CRS, particularly those arising from violations of user-specific content suitability constraints.
Existing LLM safety and alignment approaches are ill-suited to this setting~\cite{ma2026safety}. Most safety mechanisms are designed to enforce global or population-level constraints, such as generic content moderation or refusal policies, and cannot account for personalized safety requirements that vary across users and contexts. These methods lack the ability to condition safety decisions on implicit signals revealed through conversation, and therefore fail to distinguish between benign and harmful uses of the same content. As a result, they often permit recommendations that conflict with a user’s individual safety sensitivities, such as age, cultural norms, religious practices, or mental health history, leading to harmful or offensive outcomes, as shown in Figure \ref{fig:unsafe}. These failures are not isolated edge cases but reflect a fundamental mismatch between current LLM alignment objectives and the requirements of safe, personalized recommendations.
We therefore define \textbf{Personalized Safety Alignment} in CRS as the ability of a system to strictly adhere to user-specific content suitability constraints inferred from both explicit and implicit conversational signals, while preserving recommendation relevance and utility. A safe CRS must jointly reason about personalization and safety, rather than treating safety as a global or uniform constraint.

To bridge this gap, we introduce the first user-centric safety analysis benchmark for CRS, \benchmarkname{}, which augments the widely used conversational movie recommendation dataset Reddit-V2~\cite{zhu2025rank} with explicit safety annotations, establishing the movie (\benchmarkmoviename{}) and game (\benchmarkgamename{}) domains as a generalizable case study for reasoning about sensitive content in conversational recommendation. \benchmarkname{} operationalizes personalized safety through a fine-grained representation of user sensitivities. Specifically, we introduce the notion of \textit{Latent Traits}, user-specific sensitivity profiles (e.g., history of self-harm, strict aversion to sexual violence, or phobia of needles), that directly map to structured content metadata.
We further develop a trait assignment pipeline that infers latent traits from conversational context and integrates them with fine-grained content severity scores derived from DoesTheDogDie (DDD)~\cite{kovacs2025datasets} and IMDb Parent Guides~\cite{haworth2023imdb} for movies and ESRB Ratings and Content Descriptors for games. This design enables \benchmarkmoviename{} and \benchmarkgamename{} to evaluate whether CRS models can identify and respect subtle, personalized safety constraints that are typically overlooked by generic or global filtering mechanisms. Our benchmark quantifies user-centric safety in CRS by introducing a verifiable ground truth, addressing the stochastic limitations and hallucinations inherent in LLM-as-a-judge evaluations.

Building on the personalized safety challenges exposed by \benchmarkname{}, we find that existing safety alignment techniques are insufficient for LLM-based CRS. Approaches such as Reinforcement Learning from Human Feedback (RLHF) \cite{dai2023safe, ji2025pku, tan2025equilibrate} struggle to disentangle an item’s positional utility from its semantic safety, often defaulting to position-based heuristics rather than reasoning about content suitability~\cite{zheng2023judging}. Moreover, enforcing non-negotiable safety constraints frequently leads to reward signal collapse, pushing models toward unstable extremes of either unsafe recommendation or excessive refusal.
These issues are further amplified in previous group-based optimization methods such as GRPO, which are susceptible to reward hacking~\cite{zhang2025grpo, mroueh2025reinforcement}. To address these limitations, we introduce a two-stage pipeline based on Supervised Fine-Tuning and Group reward–Decoupled Normalization Policy Optimization (GDPO) \cite{liu2026gdpo}, termed as Safe-SFT and Safe-GDPO, respectively. To optimize for both safety and relevance, Safe-GDPO is used instead of GRPO. By decoupling and normalizing each reward dimension independently, GDPO prevents advantage collapse and ensures a stable signal for multi-reward optimization. Both methods are explicitly designed to balance recommendation utility and personalized safety. The resulting model, \modelname{}, demonstrates that avoiding such failures requires explicit reasoning over user-specific safety constraints. On \benchmarkmoviename{}, SafeCRS achieves near-zero violation rates across all backbones while matching or surpassing the recommendation quality of GPT-4. On \benchmarkgamename{}, SafeCRS outperforms the best baseline by $3.7\times$ in Recall@5 and $3.3\times$ in NDCG@5, demonstrating strong cross-domain generalizability.

\new{In summary, our contributions are:
(1) \textbf{Problem formalization.} We formalize personalized safety alignment in LLM-based CRS as a user-level constraint satisfaction problem, an underexplored vulnerability beyond population-level moderation.
(2) \textbf{\benchmarkname{} benchmark.} We introduce \benchmarkname{}, the first user-centric safety benchmark for CRS, with trait-conditioned risk scores over 24{,}408 movies and 9{,}722 games.
(3) \textbf{\modelname{}.} We propose \modelname{}, in which Safe-SFT couples ground-truth filtering, constraint injection, and safety reasoning supervision in one fine-tuning stage, and Safe-GDPO applies per-reward normalize-then-sum aggregation to balance sparse relevance against dense safety signals.}

\section{Related Works}

\subsection{Safety Alignment in LLMs}

The foundational paradigm for LLM safety emerged from RLHF. InstructGPT \cite{ouyang2022training} established the standard three-stage pipeline of supervised fine-tuning, reward model training, and Proximal Policy Optimization (PPO), demonstrating that alignment with human intent is critical for model safety. Subsequent works like Llama~2 \cite{touvron2023llama} refined this by utilizing dual reward models to separately optimize helpfulness and safety, supported by extensive red-teaming. To address the scalability limits of human labeling, Constitutional AI (CAI) \cite{bai2022constitutional} introduced Reinforcement Learning from AI Feedback (RLAIF), where models critique their own outputs based on a written set of principles. More recently, Direct Preference Optimization (DPO) \cite{rafailov2023direct} simplified this pipeline by optimizing the policy directly from preference data, eliminating the need for explicit reward modeling and stabilizing the alignment process.
A robust ecosystem of safety benchmarks has emerged alongside these methods. Benchmarks such as TrustLLM \cite{huang2024trustllm} and SafetyBench \cite{zhang2024safetybench} evaluate models across universal dimensions like toxicity, fairness, and privacy, while HarmBench \cite{mazeikaharmbench} standardizes red-teaming protocols. However, a critical limitation persists across these paradigms: safety is defined at the population level. Existing alignment techniques \cite{ouyang2022training,rafailov2023direct} and evaluation frameworks \cite{mazeikaharmbench} treat safety as a universal constraint, blocking objectively harmful content for all users identically. They lack the granularity to address personalized safety sensitivities, where the appropriateness of content (e.g., horror movies, non-halal food) depends entirely on the user's specific attributes, such as age, culture, or trauma history. Our work resolves this issue with a two-stage training pipeline for user-sensitive safety constraint alignment.

\subsection{LLM-based CRS}

Early CRS relied heavily on structured knowledge to bridge the semantic gap between dialogue and recommendation. Foundational works like ReDial \cite{li2018towards} introduced human-annotated datasets, while subsequent approaches such as KBRD \cite{chen2019towards}, KGSF \cite{zhou2020improving}, and UniCRS \cite{wang2022towards} utilized Knowledge Graphs (KGs) and Graph Neural Networks (GNNs) to unify user preference reasoning with natural language generation.
The advent of LLMs has transformed this landscape by enabling more flexible, training-free interaction. Methods like Chat-Rec \cite{gao2023chat} leverage LLMs as interactive agents by converting user profiles into prompts. To improve domain adaptation, InstructRec \cite{zhang2025recommendation} and TALLRec \cite{bao2023tallrec} treat recommendation as an instruction-following task, demonstrating that fine-tuning relatively small LLMs (e.g., LLaMA-7B) on personalized instructions can outperform larger generalist models.
Despite these advances in accuracy, recommendation safety remains a critical blind spot. While the field has extensively studied fairness \cite{wang2023survey} to mitigate statistical biases and ensure equal treatment across groups, these methods do not address content safety at the individual level. Existing safety metadata sources (e.g., IMDb Parental Guides, DoesTheDogDie) remain external lookup tools disconnected from the CRS inference process. Consequently, current systems lack the granularity to handle personalized safety sensitivities, such as trauma triggers, phobias, or religious restrictions, leaving users vulnerable to recommendations that are "fair" in the aggregate but harmful to the individual. To address these limitations, we propose \modelname{}, an LLM-based CRS that shifts the focus from item attributes to individual user needs. This user-centric method safeguards sensitive preferences and yields more optimized and personalized results.

\section{\benchmarkname{} Dataset and Benchmark}

\begin{figure}[t]
  \centering
  \includegraphics[width=1\columnwidth]{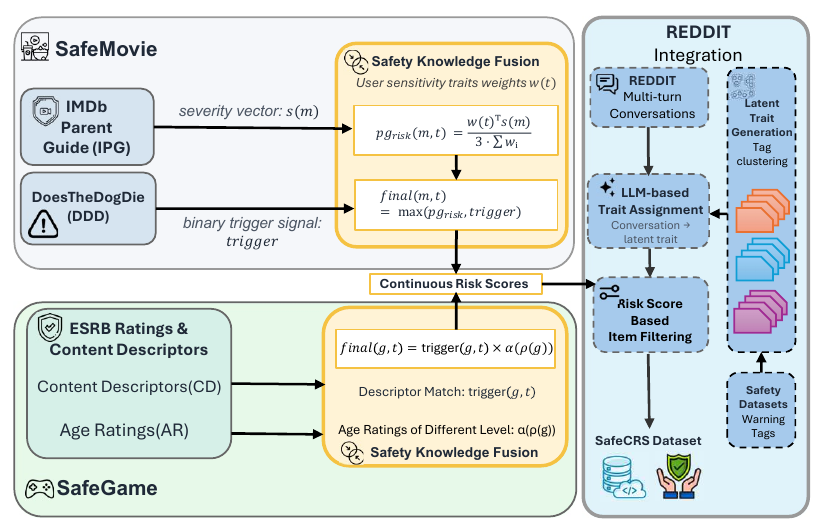}
  \vspace{-10pt}
  \caption{Overview of the \benchmarkname{} benchmark generation pipeline. We construct a ground-truth dataset for safety evaluation by integrating \benchmarkmoviename{} and \benchmarkgamename{} safety parts with conversations. The pipeline fuses domain-specific safety descriptors with user sensitivity traits extracted from Reddit, utilizing a continuous risk scoring mechanism to rigorously quantify recommendation safety.}
  \Description{A framework diagram showing the SafeCRS system: benchmark construction from DDD and IPG data sources through a Safety Knowledge Base, integration with REDDIT-V2 conversational data, and a two-stage training pipeline consisting of Safe-SFT with safety reasoning supervision and Safe-GDPO with per-rank reward decomposition.}
  \label{fig:data_pipeline}
\end{figure}

We construct the \benchmarkname{} as a multi-domain safety-aware CRS benchmark consisting of two domain datasets: \benchmarkmoviename{} (Movies) and \benchmarkgamename{} (Games).
Both domains follow a unified two-stage construction pipeline.
First, we build a domain-specific \emph{Safety Oracle} by mapping granular content ratings into trait-conditioned, computable risk scores.
Second, we perform \emph{Conversational Benchmark Integration} by aligning these item-level safety profiles with real-world recommendation conversations from Reddit using an LLM-based latent-trait inference and verification workflow.
Figure~\ref{fig:data_pipeline} illustrates the full construction process.

\subsection{\benchmarkname{} Safety Knowledge Base}
\label{sec:safety_kb}

\subsubsection{\benchmarkmoviename{} Oracle}
We construct a movie safety knowledge base by fusing DoesTheDogDie (DDD)~\cite{kovacs2025datasets} and IMDb Parent Guide (IPG)~\cite{haworth2023imdb}.
IPG provides severity annotations in \textit{None, Mild, Moderate, Severe} for five coarse dimensions:
\{Sex/Nudity, Violence/Gore, Profanity, Alcohol/Drugs/Smoking, Frightening/Intense Scenes\}.
We map them to an integer scale and represent each movie $m$ as a severity vector
\begin{equation}
    \mathbf{s}(m) = [s_{\text{sex}}, s_{\text{viol}}, s_{\text{prof}}, s_{\text{sub}}, s_{\text{fright}}] \in \{0,1,2,3\}^5,
\end{equation}
where $0/1/2/3$ correspond to None/Mild/Moderate/Severe, respectively.
DoesTheDogDie (DDD) is a community-curated safety dataset that annotates media content with 137 fine-grained warning tags (e.g., ``Is there blood/gore?'' and ``Is there sexual content?'') to support personalized trigger-aware filtering.

\paragraph{Trait taxonomy construction.}
DDD complements IPG by offering fine-grained community-curated trigger tags (over 80 categories), but many are semantically overlapping or paraphrastic.
\new{To obtain a compact yet expressive trait space, our taxonomy is grounded in the safety dimensions identified by Stray~et~al.~\cite{stray2024building} and adapted specifically to the movie and game domains.
Concretely, we collect 200+ DDD trigger labels and group semantically similar ones via LLM-guided clustering~\cite{zhang2023clusterllm},} producing a fixed set of $20$ \emph{explicit} user sensitivity traits
$\mathcal{T}=\{t_1,\ldots,t_{20}\}$ (e.g., \textit{Anti-gore}, \textit{Substance avoidance}, \textit{Kid-safety}).
\new{The full list with avoid-tags and per-IPG weights appears in Appendix~\ref{app:traits}; extending the taxonomy only requires adding a new mapping entry. We pick $20$ as a balance between granularity and practicality: the distribution is already long-tailed (Table~\ref{tab:safemovie_trait_dist}), and finer splits would induce severe data sparsity.}

\paragraph{Trait-to-IPG weights $\mathbf{w}(t)$.}
Each trait $t\in\mathcal{T}$ is associated with a non-negative weight vector $\mathbf{w}(t)\in\mathbb{R}_{\ge 0}^{5}$ that measures how strongly each IPG dimension contributes to that trait.
We construct $\mathbf{w}(t)$ using a lightweight, reproducible procedure:
(i) for each trait, we prompt an LLM to map its underlying DDD cluster labels to the most relevant IPG dimension(s);
(ii) we aggregate these mappings into a frequency-based relevance profile over the five IPG dimensions and normalize it to obtain $\mathbf{w}(t)$;
and (iii) we conduct a brief manual sanity check to correct rare mis-assignments.
This yields interpretable, non-negative weights and avoids learning an unconstrained mapping from scarce supervision.

\paragraph{Continuous parental-guidance risk.}
Given a movie $m$ and trait $t$, we compute a continuous risk score as a normalized weighted sum:
\begin{equation}
    \textit{pg\_risk}(m,t) = \frac{\mathbf{w}(t)^\top \mathbf{s}(m)}{3\sum_i w_i(t)} \in [0,1].
\end{equation}
The constant $3$ appears because each IPG dimension has maximum severity $3$; dividing by $3\sum_i w_i(t)$ normalizes the weighted sum to the unit interval, making risk scores comparable across traits.

\paragraph{Hard triggers beyond coarse IPG categories.}
While IPG offers broad severity dimensions, it may miss highly specific triggers important for personalized safety.
For example, a user may wish to avoid \emph{animal death}, \emph{needles/medical procedures}, \emph{self-harm/suicide}, or \emph{sexual assault}, which are explicitly captured by DDD but not uniquely determined by the five coarse IPG dimensions.
To enforce such specific triggers, we incorporate a binary trigger signal $\textit{trigger}(m,t)\in\{0,1\}$ from DDD and define the final trait-conditioned risk as
\begin{equation}
    \textit{final\_risk}(m,t) = \max\!\big(\textit{pg\_risk}(m,t), \textit{trigger}(m,t)\big).
\end{equation}
Here, $\textit{trigger}(m,t)=1$ acts as a hard veto, assigning maximal risk regardless of the IPG severity profile.

\subsubsection{\benchmarkgamename{} Oracle}
\label{sec:game_oracle}
We construct a parallel safety knowledge base for games grounded in the Entertainment Software Rating Board (ESRB).
ESRB provides (i) a categorical \emph{age rating} $\rho(g)\in\{\text{E},\text{E10+},\text{T},\text{M},\text{AO}\}$ and (ii) a set of \emph{content descriptors} (e.g., ``Blood and Gore'', ``Intense Violence'', ``Strong Language'').
We define a set of $10$ game-domain sensitivity traits $\mathcal{T}_g=\{t_1,\dots,t_{10}\}$ grounded in ESRB descriptors (e.g., \textit{Anti-gore}, \textit{Extreme Violence}, \textit{Sexual Content}, \textit{Substance Avoidance}, \textit{Profanity}, \textit{Kid-safety}, \textit{Horror Avoidance}). The complete list is in Appendix \ref{app:gametraits}.

For a given game $g$ and trait $t\in\mathcal{T}_g$, we first compute a binary trigger signal based on descriptor matching:
\begin{equation}
    \textit{trigger}(g,t)=
    \begin{cases}
    1, & \text{if } \mathcal{D}(t)\cap \mathcal{D}(g)\neq \emptyset,\\
    0, & \text{otherwise},
    \end{cases}
\end{equation}
where $\mathcal{D}(t)$ denotes the set of ESRB descriptors mapped to trait $t$ and $\mathcal{D}(g)$ denotes the descriptors assigned to game $g$.
We then modulate the trigger by a rating-based weight $\alpha(\rho(g))\in[0.2,1.0]$:
\begin{equation}
    \alpha(\rho)=
    \begin{cases}
    0.2, & \rho=\text{E},\\
    0.4, & \rho=\text{E10+},\\
    0.6, & \rho=\text{T},\\
    0.8, & \rho=\text{M},\\
    1.0, & \rho=\text{AO},
    \end{cases}
\end{equation}
where E, E10+, T, M, and AO are ESRB age-rating categories:
E = \textit{Everyone},
E10+ = \textit{Everyone 10+},
T = \textit{Teen},
M = \textit{Mature 17+},
and AO = \textit{Adults Only 18+}
and define the final risk score as
\begin{equation}
    \textit{final\_risk}(g,t)=\textit{trigger}(g,t)\cdot \alpha\!\left(\rho(g)\right).
\end{equation}
A game is considered unsafe for a given trait when its risk score exceeds a threshold $\tau$.

\subsection{Conversational Benchmark Integration}
\label{sec:conv_integration}
At this stage, we have domain-specific Safety Oracles that return trait-conditioned risk scores for items.
We next integrate these item-level safety profiles with real-world conversational recommendation data to build safety-aware conversational benchmarks for both domains.

\subsubsection{{\benchmarkmoviename{}}: Reddit-V2 integration.} We ground \benchmarkmoviename{} in Reddit-V2~\cite{zhu2025rank}, which contains multi-turn recommendation conversations paired with ground-truth movie lists.
While Reddit-V2 provides realistic user intents and conversational context, it does not explicitly annotate user-specific safety constraints.

\paragraph{Latent traits inference.} Given a conversation context, we employ an LLM to infer the user’s safety preferences as natural-language \emph{latent trait statements} aligned with our explicit trait taxonomy (e.g., ``Please avoid movies with excessive blood and gore.'' for \textit{Anti-gore / squeamish}). (Appendix \ref{app:redditv2_example_movie})

\paragraph{Safety alignment.} We map each referenced movie to the Movie Oracle and compute $\textit{final\_risk}(m,t)$ for the inferred traits.
Items whose risk exceeds a fixed threshold $\tau$ are treated as safety-violating under the conversation-implied constraints, enabling safety-aware evaluation and training targets grounded in structured evidence.

\subsubsection{\benchmarkgamename{}: r/gamingsuggestions integration.} We ground the game benchmark, \benchmarkgamename{}, in conversations collected from the Reddit community r/gamingsuggestions, which provides authentic game recommendation requests and responses in a conversational format.
As in the movie domain, these conversations do not explicitly label safety constraints.

\paragraph{Latent trait inference.} We apply the same LLM-based inference procedure to convert each conversation into latent trait statements aligned with the game-domain trait taxonomy $\mathcal{T}_g$ (e.g., ``Avoid intense violence and gore.'' for \textit{Anti-gore} / \textit{Extreme Violence}). Details can be found in Appendix \ref{app:latent_trait_generation}.

\paragraph{Safety alignment.} We link recommended games to the Game Oracle and compute $\textit{final\_risk}(g,t)$ based on ESRB descriptors and ratings.
Games whose risk exceeds $\tau$ are considered unsafe under the inferred traits, yielding safety-aware ground truth and enabling consistent evaluation across domains.

Our safety labels are produced by deterministic, reusable oracles: once an item is linked to its structured content profile (IPG/DDD for movies; ESRB for games), trait-conditioned risks follow from simple lookups and closed-form scoring.
This is substantially more efficient and consistent than applying an LLM-as-a-Judge to score safety for every (conversation, recommendation) pair, which requires per-item prompting and may introduce variance across runs.

\section{\modelname{}}
\begin{figure*}[t]
  \centering
  \includegraphics[width=\textwidth]{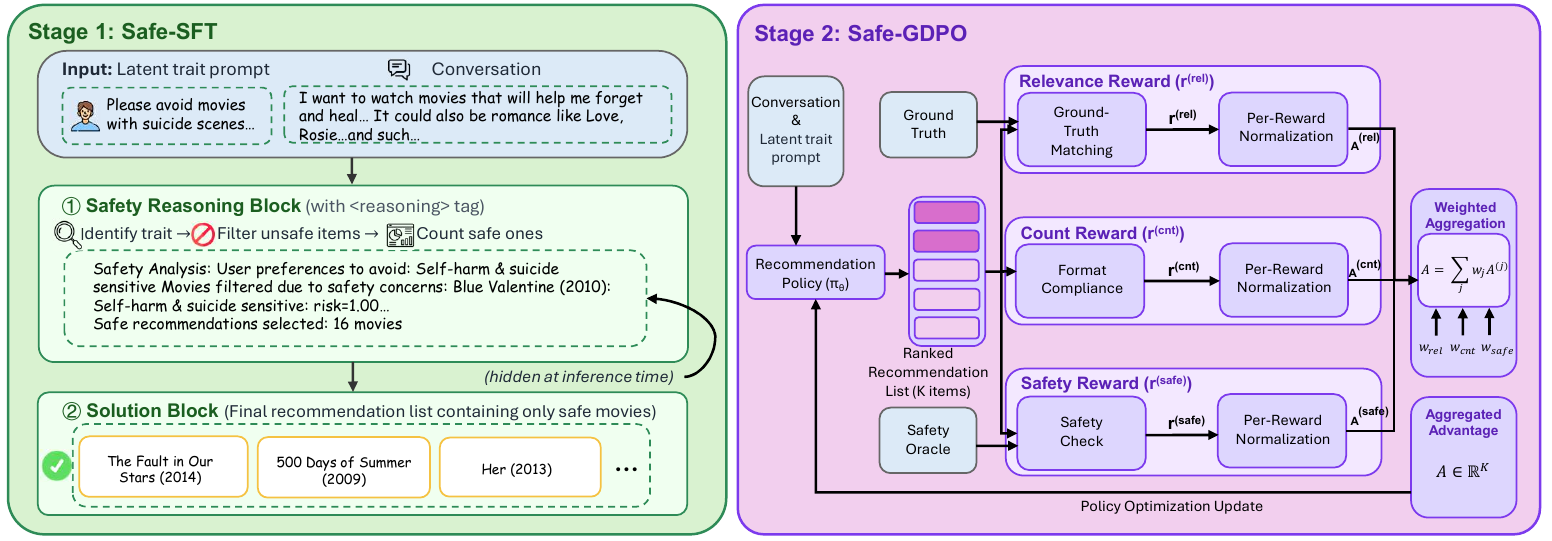}
  \caption{Two-stage training pipeline. Stage~1 (Safe-SFT) trains the model to produce a safety reasoning block that identifies and filters unsafe items, followed by a safe recommendation list. Stage~2 (Safe-GDPO) samples multiple ranked completions and applies per-rank reward decomposition, combining relevance rewards with safety penalties from a plug-in Safety Oracle, to update the policy via group-normalized advantages.}
  \Description{A framework diagram showing the SafeCRS training system: a two-stage training pipeline consisting of Safe-SFT with safety reasoning supervision and Safe-GDPO with per-rank reward decomposition. }
  \label{fig:framework}
\end{figure*}

\modelname{} is trained in two stages. Safe-SFT supervises the model to perform preference-conditioned safety filtering and to generate a final list containing only safe items. Safe-GDPO subsequently refines rank-wise recommendations under the same constraints via GDPO, ensuring sparse relevance signals remain effective alongside denser safety and format rewards. An overview of the full framework is shown in Figure~\ref{fig:framework}.

\subsection{Safe-SFT}
\label{sec:safe-sft}
We construct a supervised fine-tuning (SFT) dataset that explicitly teaches the model to (i) perform a safety analysis of candidate items and (ii) produce a final recommendation list that excludes items deemed unsafe with respect to a user-specified safety preference.
Concretely, we start from base recommendation completions (e.g., top-$N$ movie lists) produced by a general-purpose assistant (GPT-4o).
For each completion, we use an external safety database as described in Section~\ref{sec:safety_kb} that scores each candidate item under a fixed set of latent traits.
Items with risk scores exceeding a threshold $\tau$ are marked unsafe and removed.

To enable the CRS to learn the safety behavior, we structure each training target into a two-part structure:
(1) a \emph{safe reasoning} block that documents the detected preference, lists filtered items together with their safety rationale and risk scores, and summarizes the number of safe items retained; and
(2) a \emph{solution} block that contains only the final safe recommendations.
We keep the reasoning concise and strictly grounded in the classifier outputs so that the model learns to justify removals without hallucinating additional harms.

\subsection{Safe-GDPO}
\label{sec:method_safe_gdpo}

The second stage, \textbf{Safe-GDPO}, further updates the recommendation policy to improve recommendation quality while preserving user-aware safety. We address a practical challenge in safety-aware recommendation: the sparsity levels of different reward signals can vary substantially. Since recommendation tasks are always open-ended questions and human recommendation can only include part of correct answers, ground-truth matches are scarce, making the relevance reward extremely sparse, while safety and format rewards are comparatively denser.
To mitigate reward-sparsity imbalance and avoid collapsing optimization toward dense rewards, we adopt GDPO~\cite{liu2026gdpo}, which performs \emph{per-reward} normalization before aggregation.

We propose to optimize the model with three independent reward functions, including relevance, safety, and output-length compliance:
$\mathcal{R}=\{r^{(\mathrm{rel})}, r^{(\mathrm{safe})}, r^{(\mathrm{cnt})}\}$.
Each reward function produces a \emph{rank-wise} score vector over a recommendation list of length $K$, i.e.,
$\mathbf{r}^{(j)}=[r^{(j)}_1,\ldots,r^{(j)}_K]\in\mathbb{R}^K$,
where $r^{(j)}_k$ denotes the reward assigned to the item at rank $k$ under reward type $j$.

\subsubsection{Reward Design}
\paragraph{Relevance reward (binary hits).}
We compute a binary hit at each rank $k$ based on whether the recommended item matches the ground-truth movie.
Specifically, the relevance reward is defined as an indicator function that returns $1$ if the title matches the
ground truth and the predicted release year is within a tolerance of two years, and $0$ otherwise. This produces a sparse binary vector $\mathbf{r}^{(\mathrm{rel})}\in\{0,1\}^K$ over the ranked list.

\paragraph{Safety reward (rank-discounted penalties)}
A safety oracle checks whether a recommended item violates any user-specific avoid constraints.
Let $v_k\in\{0,1\}$ indicate whether the item at rank $k$ is unsafe (i.e., violates at least one constraint).
We apply a logarithmic rank discount because higher-ranked recommendations are more likely to be noticed and acted upon by the user, and thus unsafe items appearing earlier should be penalized more strongly than those appearing later:
\begin{equation}
d_k = \frac{1}{\log_2(k+1)} \text{ or }\frac{1}{\log_2(k+2)} \text{ under 0-based indexing},
\label{eq:discount}
\end{equation}
and define the safety reward as
\begin{equation}
r^{(\mathrm{safe})}_k
=
-\lambda_{\mathrm{safe}} \, v_k \, d_k,
\label{eq:safety_reward}
\end{equation}
where $\lambda_{\mathrm{safe}}>0$ controls the overall penalty magnitude, $v_k \in \{0,1\}$ is the binary violation indicator of the item at rank $k$ under the current dialogue's trait-conditioned constraints returned by the \benchmarkname{} Safety Oracle as described in Section~\ref{sec:safety_kb}, and $d_k$ is the rank discount factor in Eq.~\eqref{eq:discount}.
This formulation assigns zero penalty to safe items ($v_k=0$) and a rank-discounted negative penalty to unsafe items ($v_k=1$), with larger penalties for higher-ranked positions due to $d_k$.

\paragraph{Count reward (list-format compliance)}
Let $\hat{c}$ be the parsed number of generated recommendations and $c^\star$ be the target (we use $c^\star=10$).
We define a scalar count reward:
\begin{equation}
R^{(\mathrm{cnt})}=
\begin{cases}
+\lambda_{\mathrm{cnt}}, & \hat{c}=c^\star,\\[4pt]
-\lambda_{\mathrm{cnt}}\cdot \dfrac{|\hat{c}-c^\star|}{c^\star}, & \text{otherwise},
\end{cases}
\label{eq:count_reward_scalar}
\end{equation}
and broadcast it to all ranks:
\begin{equation}
r^{(\mathrm{cnt})}_k = R^{(\mathrm{cnt})}, \qquad \forall k\in\{1,\dots,K\}.
\label{eq:count_reward_broadcast}
\end{equation}

This reward encourages the model to output exactly $c^\star$ recommendations: it gives a positive bonus when $\hat{c}=c^\star$, and otherwise applies a linear penalty proportional to the relative deviation $|\hat{c}-c^\star|/c^\star$.
We broadcast the same scalar reward to all ranks since list-length compliance is a global property of the entire output.

\subsubsection{GDPO advantage aggregation and policy optimization}
The key component of Safe-GDPO is \emph{per-reward} normalization before combining multiple reward channels.
For each reward function $\mathbf{r}^{(j)}$, $j\in\{\mathrm{rel},\mathrm{safe},\mathrm{cnt}\}$, we compute a normalized advantage vector:
\begin{equation}
\mathbf{A}^{(j)}
=
\frac{\mathbf{r}^{(j)} - \mu\!\left(\mathbf{r}^{(j)}\right)}
{\sigma\!\left(\mathbf{r}^{(j)}\right)+\epsilon},
\label{eq:gdpo_norm}
\end{equation}
where $\mu(\cdot)$ and $\sigma(\cdot)$ denote the mean and standard deviation (computed over the batch and ranks), and $\epsilon$ is a small constant.
We then aggregate the per-reward advantages via a weighted sum:
\begin{equation}
\mathbf{A}
=
\sum_{j\in\{\mathrm{rel},\mathrm{safe},\mathrm{cnt}\}}
w_j \, \mathbf{A}^{(j)}
\in \mathbb{R}^{K},
\label{eq:gdpo_weighted_sum}
\end{equation}
where $w_j$ is a scalar weight controlling the relative contribution of reward channel $j$.
Finally, we plug the aggregated rank-wise advantage $\mathbf{A}$ into the policy optimization objective to update the model parameters.
By normalizing each reward channel separately before aggregation, GDPO preserves informative gradient signals from sparse rewards and prevents them from being overwhelmed by denser safety and format rewards.

\section{Experiment}
We design our experiments to answer the following research questions:
\textbf{RQ1}: How do existing CRS methods respect user-specific safety preferences?
\textbf{RQ2}: How effective is each training stage (Safe-SFT, Safe-GDPO) of the \modelname{} pipeline in improving CRS personalized safety?
\textbf{RQ3}: Can \modelname{} be instantiated with a unified training recipe on peer domains with different safety taxonomies?
\new{\textbf{RQ4}: Do \modelname{}'s safety improvements, measured by the oracle, align with human judgments of recommendation safety?}

\subsection{Experimental Setup}
\begin{table*}[t]
\centering
\caption{Main results on the \benchmarkmoviename{} (left) and \benchmarkgamename{} (right) benchmarks. $\uparrow$: higher is better; $\downarrow$: lower is better. Best results are in \textbf{bold} and second-best are \underline{underlined}. KBRD and NBCRS are closed-vocabulary models restricted to fixed movie catalogs and cannot generalize to the game domain. The left part of the table reports results on \benchmarkmoviename{}, while the right part reports results on \benchmarkgamename{}.}
\label{tab:main_results}
\resizebox{\textwidth}{!}{%
\begin{tabular}{l cccc cccc | cccc cccc}
\toprule
\multirow{2}{*}{\textbf{Method}} 
& \multicolumn{2}{c}{\textbf{Recall} $\uparrow$} 
& \multicolumn{2}{c}{\textbf{NDCG} $\uparrow$} 
& \multicolumn{2}{c}{\textbf{SVR} $\downarrow$} 
& \multicolumn{2}{c}{\textbf{S-DCG} $\downarrow$}
& \multicolumn{2}{c}{\textbf{Recall} $\uparrow$} 
& \multicolumn{2}{c}{\textbf{NDCG} $\uparrow$} 
& \multicolumn{2}{c}{\textbf{SVR} $\downarrow$} 
& \multicolumn{2}{c}{\textbf{S-DCG} $\downarrow$} \\
\cmidrule(lr){2-3} \cmidrule(lr){4-5} \cmidrule(lr){6-7} \cmidrule(lr){8-9}
\cmidrule(lr){10-11} \cmidrule(lr){12-13} \cmidrule(lr){14-15} \cmidrule(lr){16-17}
& @5 & @10 & @5 & @10 & @5 & @10 & @5 & @10
& @5 & @10 & @5 & @10 & @5 & @10 & @5 & @10 \\
\midrule

\multicolumn{17}{l}{\textit{Traditional CRS}} \\
KBRD              & .0115 & .0135 & .0090 & .0097 & .3752 & .3756 & 1.11 & 1.62  & - & - & - & - & - & - & - & - \\
NBCRS             & .0451 & .0766 & .0312 & .0420 & .5380 & .5083 & 1.64 & 2.40  & - & - & - & - & - & - & - & - \\

\midrule
\multicolumn{17}{l}{\textit{CRAG (Collaborative Retrieval + Closed-source LLM)}} \\
CRAG (GPT-4o)     & \underline{.0826} & .1217 & .0581 & .0715 & .4007 & .3894 & 1.20 & 1.80  & .0422 & .0422 & .0833 & .0833 & .0664 & .0655 & 0.20 & 0.26 \\
CRAG (Haiku4.5)   & .0668 & .0972 & .0476 & .0577 & .3587 & .3524 & 1.06 & 1.61  & .0162 & .0164 & .0354 & .0361 & .0367 & .0365 & 0.09 & 0.11 \\
CRAG (Gemini2.0)  & .0759 & \underline{.1230} & .0563 & .0720 & .3695 & .3690 & 1.09 & 1.68  & .0036 & .0036 & .0022 & .0022 & \textbf{.0017} & \underline{.0017} & \underline{0.02} & \textbf{0.02} \\

\midrule
\multicolumn{17}{l}{\textit{Closed-source LLMs (Zero-shot)}} \\
GPT-4             & .0705 & .1128 & .0517 & .0664 & .4454 & .4359 & 1.31 & 2.00  & .0363 & .0363 & .0754 & .0754 & .0630 & .0679 & 0.17 & 0.27 \\
GPT-5.2           & \textbf{.0827} & \textbf{.1379} & \textbf{.0627} & \textbf{.0815} & .3508 & .3369 & 1.03 & 1.55  & .0422 & .0426 & .0833 & .0827 & .0400 & .0380 & 0.12 & 0.17 \\

\midrule
\multicolumn{17}{l}{\textit{Open-source LLMs (Zero-shot)}} \\
Gemma3-27B        & .0676 & .0973 & .0534 & .0636 & .3548 & .3684 & 0.99 & 1.61  & .0081 & .0081 & .0117 & .0117 & \underline{.0018} & \textbf{.0014} & \textbf{0.00} & \underline{0.01} \\
Llama3-SDSC-70B   & .0650 & .0879 & .0483 & .0564 & .3642 & .3443 & 1.09 & 1.61  & .0735 & .0740 & .1031 & .1034 & .0515 & .0521 & 0.14 & 0.20 \\

\midrule
\multicolumn{17}{l}{\textit{SafeCRS}} \\
Qwen2.5-0.5B      & .0732 & .0922 & .0547 & .0597 & \textbf{.0011} & \textbf{.0006} & 0.32 & 0.35  
                  & .1542 & .1789 & .1884 & .1902 & .0774 & .0792 & 0.11 & 0.12 \\
Qwen3-8B          & .0743 & .0951 & .0597 & .0614 & \underline{.0013} & \underline{.0022} & \textbf{0.039} & \textbf{0.043}
                  & \underline{.2345} & \textbf{.2996} & \underline{.3108} & \underline{.3269} & .0232 & .0176 & 0.03 & 0.04 \\
Llama-3.2-3B      & .0711 & .0996 & .0605 & .0592 & .0178 & .0101 & 0.07 & 0.12
                  & .2276 & .2452 & .2939 & .3002 & .0547 & .0583 & 0.07 & 0.09 \\
Llama-3.1-8B      & .0774 & .1111 & \underline{.0625} & \underline{.0737} & .0122 & .0087 & \underline{0.039} & \underline{0.048}
                  & \textbf{.2702} & \underline{.2907} & \textbf{.3404} & \textbf{.3614} & .0189 & .0132 & 0.04 & 0.04 \\
\bottomrule
\end{tabular}%
}
\end{table*}
\textbf{Datasets.}
We conduct all experiments on \benchmarkmoviename{} and \benchmarkgamename{}. \benchmarkmoviename{} contains 19,086/1,127/1,212 (train/val/test) samples, each consisting of a multi-turn conversational context, ground-truth movie recommendations, and a user sensitivity trait from 20 predefined traits (Appendix \ref{app:safemovie_traits}). The safety knowledge base covers 24,408 movies with fine-grained trait sensitivity scores. \benchmarkgamename{} contains 10,257/1,282/1,283 (train/val/test) samples covering 1,403 unique games from a catalog of 9,722, where risk scores are computed by multiplying a binary content-descriptor trigger with an ESRB rating-based weight ($0.2$--$1.0$) and thresholded at $\tau = 0.66$.

\noindent\textbf{Implementation Details.}
For Safe-SFT, we fine-tune each base model using the TRL SFTTrainer with a learning rate of $5\times10^{-5}$, cosine scheduling with 5\% warmup, and a maximum sequence length of 1,024 tokens. Training runs for 10 epochs with a per-device batch size of 12 and gradient accumulation of 8 steps.
For Safe-GDPO, we initialize from the Safe-SFT checkpoint and train for 2 epochs with a learning rate of $10^{-6}$, KL penalty $\beta = 10^{-3}$, and 8 sampled completions per prompt. The safety penalty uses $\lambda_{\text{safe}} = 1.0$, $P_{\text{safe}} = 1.0$, and a risk threshold $\tau = 0.66$. All models are trained in bfloat16 precision with paged AdamW 8-bit optimizer.

\subsection{Baselines}

We compare \modelname{} with four categories of baselines:

\noindent\textbf{Traditional CRS.}
\textbf{KBRD}~\cite{chen2019towards} integrates knowledge graphs (DBpedia) into conversational recommendation through relational graph convolutional networks.
\textbf{NBCRS}~\cite{xie2024neighborhood} is a neural baseline CRS trained on the Reddit movie recommendation corpus.
The output spaces of these two models are restricted to fixed movie catalogs derived from their training corpora (ReDial and Reddit, respectively). Unlike LLM-based baselines, they cannot generalize to new domains without retraining from scratch, and are therefore evaluated only on \benchmarkmoviename{}.

\noindent\textbf{CRAG (Collaborative Retrieval + LLMs).}
\textbf{CRAG}~\cite{10.1145/3696410.3714908} is a retrieval-augmented framework that enhances LLM-based CRS by incorporating collaborative filtering signals. We evaluate CRAG with three LLM backends: GPT-4o, Claude Haiku~4.5, and Gemini~2.0.

\noindent\textbf{Closed-source LLMs (Zero-shot).}
We evaluate two proprietary models in a zero-shot prompting setup: \textbf{GPT-4}~\cite{achiam2023gpt} and \textbf{GPT-5.2}~\cite{singh2025openai} are prompted directly to generate recommendations from conversational context.

\noindent\textbf{Open-source LLMs (Zero-shot).}
We include zero-shot results for \textbf{Gemma3-27B}~\cite{team2025gemma} and \textbf{Llama3-SDSC-70B}~\cite{dubey2024llama} as additional baselines. The zero-shot performance of the four backbone models (\textbf{Qwen2.5-0.5B}~\cite{qwen2025qwen25technicalreport}, \textbf{Qwen3-8B}~\cite{yang2025qwen3}, \textbf{Llama-3.2-3B}~\cite{dubey2024llama}, \textbf{Llama-3.1-8B}~\cite{dubey2024llama}) is reported in the ablation tables.

Notably, none of the above baselines incorporate any safety-awareness mechanism. This gap underscores the need for a dedicated safety-aware training framework.

\subsection{Metrics}
We evaluate models along two orthogonal axes: \emph{recommendation quality} and \emph{safety compliance}. Let $\hat{\mathcal{R}} = (\hat{r}_1,\dots,\hat{r}_K)$ denote the top-$K$ recommendation list produced by a model, $\mathcal{G}$ the set of ground-truth items, and $t$ the user's assigned sensitivity trait with risk threshold $\tau$.

\subsubsection{Recommendation Quality}
To evaluate recommendation quality, we employ two standard top-$K$ metrics: \textbf{Recall@$K$}, which measures the fraction of ground-truth items retrieved within the top $K$ candidates, and \textbf{NDCG@$K$}, which further accounts for ranking position by rewarding relevant items placed higher in the list.

\subsubsection{Safety Compliance}
To quantify the degree to which a model respects user-specific safety constraints, we define a per-item violation indicator:
\begin{equation}
    v_k = \mathbbm{1}\!\left[\text{final}(\hat{r}_k, t) \geq \tau\right],
\end{equation}
where $\text{final}(\cdot)$ is the final risk score from Section~3.1 and $\tau = 0.66$ corresponds to approximately \emph{Moderate} severity. We then report two complementary metrics (both \emph{lower is better}):

\noindent\textbf{Safety Violation Ratio (SVR@$K$)} captures the proportion of recommendations that violate the user's safety constraint:
\begin{equation}
    \text{SVR@}K = \frac{1}{K}\sum_{k=1}^{K} v_k.
\end{equation}

\noindent\textbf{Sensitivity DCG (S-DCG@$K$)} penalizes violations at higher ranks more heavily, mirroring the position-aware logic of NDCG:
\begin{equation}
    \text{S\text{-}DCG@}K = \sum_{k=1}^{K}\frac{v_k}{\log_2(k+1)}.
\end{equation}
S-DCG reflects the fact that unsafe items ranked higher are more harmful, as users are more likely to engage with top-ranked recommendations.

Together, Recall/NDCG and SVR/S-DCG allow us to diagnose the safety--relevance trade-off: an ideal model maximizes Recall and NDCG while minimizing SVR and S-DCG.

\subsection{Main Results (RQ1 and RQ3)}

Table~\ref{tab:main_results} presents the main results on the \benchmarkname{} benchmark.

\begin{figure*}[t]
  \centering
  \includegraphics[width=\textwidth]{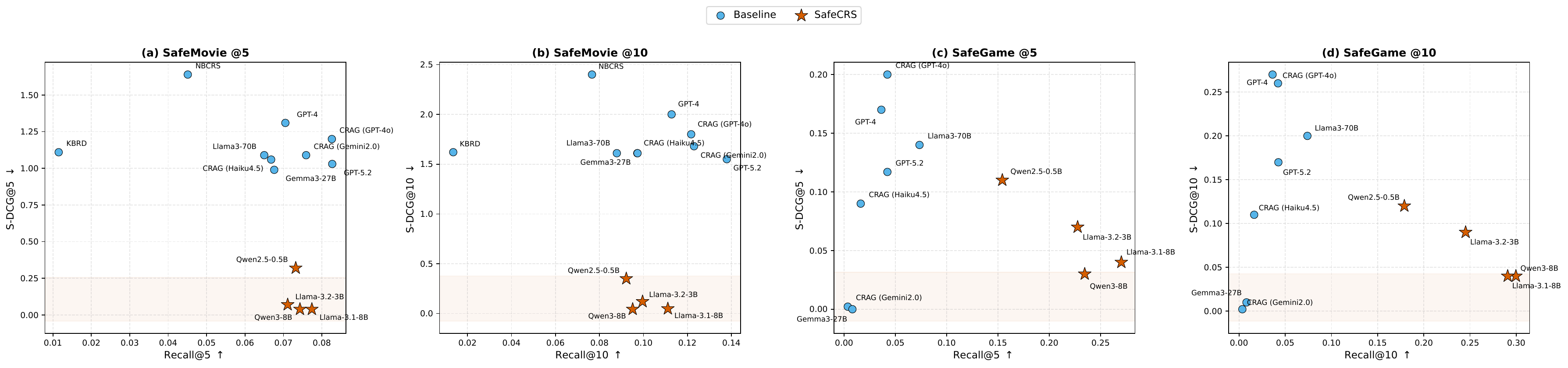}
  \caption{Safety--relevance trade-off across all methods on \benchmarkname{}}
  \label{fig:tradeoff}
\end{figure*}

Figure~\ref{fig:tradeoff} visualizes the safety--relevance trade-off by plotting Recall@$K$ against S-DCG@$K$ for each method.

None of the baselines are designed to respect user-specific safety constraints, and this is reflected in universally high violation rates. Among closed-source LLMs, GPT-4 exhibits the highest SVR (SVR@5 = 0.4454) despite strong recommendation quality, demonstrating that \emph{better recommendation capabilities do not translate to better safety}. Even GPT-5.2, the strongest baseline in recommendation quality (Recall@10 = 0.1379, NDCG@10 = 0.0815), still violates user constraints in top-ranked movies (SVR@5 = 0.3508). CRAG improves recommendation grounding through collaborative retrieval, yet its safety violations remain comparable to or worse than zero-shot LLMs (e.g., CRAG with GPT-4o: SVR@5 = 0.4007; with Gemini~2.0: SVR@5 = 0.3695), confirming that retrieval augmentation alone does not address personalized safety. Traditional CRS methods perform no better: NBCRS exhibits the worst safety performance (SVR@5 = 0.5380). These results confirm that no existing method---regardless of model scale, architecture, or retrieval augmentation---addresses personalized safety in conversational recommendation.
This pattern is clearly visible in Figure~\ref{fig:tradeoff}, where all baselines cluster in the upper-left region (high S-DCG, low-to-moderate Recall) across all four evaluation settings, indicating that neither scaling model capacity nor incorporating retrieval augmentation meaningfully reduces position-weighted safety violations.

\noindent \textbf{\modelname{} achieves strong safety with competitive relevance.}
In contrast, all \modelname{} variants drastically reduce safety violations while maintaining competitive recommendation quality.
On \benchmarkmoviename{}, Llama-3.1-8B with \modelname{} achieves Recall@10 = 0.1111 and NDCG@10 = 0.0737, comparable to GPT-5.2 (0.1379 / 0.0815), while reducing SVR@5 from 0.3508 to 0.0122---a $96.5\%$ relative reduction.
Even the smallest backbone, Qwen2.5-0.5B, achieves near-zero violation rates (SVR@5 = 0.0011) with Recall@5 = 0.0732, surpassing GPT-4's recommendation quality (Recall@5 = 0.0705) while virtually eliminating safety violations.
In Figure~\ref{fig:tradeoff}, \modelname{} variants consistently occupy the bottom-right region across all four panels, confirming that the two-stage training pipeline shifts models toward the Pareto frontier where both safety and relevance are simultaneously optimized.

\noindent \textbf{\benchmarkgamename{} results mirror the movie-domain findings.}
We observe the same pattern on \benchmarkgamename{}: existing baselines are not designed to satisfy user-specific safety constraints, resulting in non-trivial violation rates, while \modelname{} achieves a strong safety--relevance trade-off.
For example, after the full \modelname{} pipeline, the 8B backbones attain low violation rates while delivering substantially higher recommendation quality than non-safety-aware baselines (e.g., Llama-3.1-8B: SVR@5 = 0.0189 with Recall@10 = 0.2907; Qwen3-8B: SVR@5 = 0.0232 with Recall@10 = 0.2996).
We further note that some baselines may appear to have very low SVR/S-DCG on \benchmarkgamename{}, but this can be an artifact of extremely low catalog match rates (i.e., very low Recall), where failing to recommend in-catalog items trivially reduces measurable violations rather than reflecting genuine safety awareness.

Overall, these results confirm that \modelname{} improves personalized safety without sacrificing relevance across both datasets from different domains.

\subsection{Ablation Study (RQ2)}

To analyze the contribution of each training stage, we test the performance progression from the zero-shot baseline through Safe-SFT to the full \modelname{} pipeline (Safe-SFT + Safe-GDPO) for each backbone model on the \benchmarkname{} benchmark.

\begin{table}[t]
\centering
\caption{Ablation study on \benchmarkmoviename{}{}. For each backbone model, we compare the zero-shot baseline, Safe-SFT only, and the full \modelname{} pipeline (Safe-SFT + Safe-GDPO). Metrics follow the same notation as Table~\ref{tab:main_results}.}
\label{tab:ablation}
\resizebox{\columnwidth}{!}{
\begin{tabular}{ll cccc cccc}
\toprule
\multirow{2}{*}{\textbf{Backbone}} & \multirow{2}{*}{\textbf{Training Stage}} & \multicolumn{2}{c}{\textbf{Recall} $\uparrow$} & \multicolumn{2}{c}{\textbf{NDCG} $\uparrow$} & \multicolumn{2}{c}{\textbf{SVR} $\downarrow$} & \multicolumn{2}{c}{\textbf{S-DCG} $\downarrow$} \\
\cmidrule(lr){3-4} \cmidrule(lr){5-6} \cmidrule(lr){7-8} \cmidrule(lr){9-10}
 & & @5 & @10 & @5 & @10 & @5 & @10 & @5 & @10 \\
\midrule
\multirow{3}{*}{Qwen2.5-0.5B}
 & Zero-shot    & .0008 & .0008 & .0005 & .0005 & .0562 & .0503 & 0.14 & 0.16 \\
 & + Safe-SFT   & .0467 & .0732 & .0333 & .0422 & .0177 & .0133 & 0.54 & 0.68 \\
 & + Safe-GDPO  & .0732 & .0922 & .0547 & .0597 & .0011 & .0006 & 0.32 & 0.35 \\
\midrule
\multirow{3}{*}{Qwen3-8B}
 & Zero-shot    & .0182 & .0265 & .0127 & .0154 & .1253 & .1135 & 0.32 & 0.46 \\
 & + Safe-SFT   & .0687 & .0918 & .0506 & .0588 & .0135 & .0089 & 0.043 & 0.049 \\
 & + Safe-GDPO  & .0743 & .0951 & .0597 & .0614 & .0013 & .0022 & 0.039 & 0.043 \\
\midrule
\multirow{3}{*}{Llama-3.2-3B}
 & Zero-shot    & .0529 & .0682 & .0404 & .0454 & .2261 & .1967 & 0.68 & 0.95 \\
 & + Safe-SFT   & .0687 & .0804 & .0513 & .0550 & .0337 & .0369 & 0.09 & 0.12 \\
 & + Safe-GDPO  & .0711 & .0996 & .0605 & .0592 & .0178 & .0101 & 0.07 & 0.12 \\
\midrule
\multirow{3}{*}{Llama-3.1-8B}
 & Zero-shot    & .0505 & .0676 & .0368 & .0422 & .2177 & .1886 & 0.67 & 0.93 \\
 & + Safe-SFT   & .0744 & .1059 & .0613 & .0716 & .0224 & .0137 & 0.067 & 0.076 \\
 & + Safe-GDPO  & .0774 & .1111 & .0625 & .0737 & .0122 & .0087 & 0.039 & 0.048 \\
\bottomrule
\end{tabular}
}
\end{table}

Table~\ref{tab:ablation} reports the ablation on the \benchmarkmoviename{} benchmark; Table~\ref{tab:safegames_ablation} reports the same stage-wise ablation on \benchmarkgamename{}.



\new{\subsection{Human Evaluation (RQ4)}\label{sec:humaneval}}

\new{To complement our oracle-based evaluation, which is grounded in aggregated human judgements from DDD, IMDb Parent Guide, and ESRB, we conducted two targeted human studies on \benchmarkmoviename{} that directly assess (i) the reliability of the \benchmarkname{} construction pipeline and (ii) the safety of \modelname{}'s recommendations as judged by human raters.}

\new{\paragraph{\benchmarkname{} construction audit ($n=50$).}
We randomly sampled $50$ \benchmarkmoviename{} test cases and asked annotators to score three properties of each constructed sample on a per-instance basis:
(a)~the \emph{constraint naturalness} of the injected latent-trait statement (1--5 Likert);
(b)~whether the retained ground-truth recommendations \emph{respect the injected constraint};
(c)~whether the LLM-assigned explicit trait \emph{fits the user's conversational context}.
Results are summarized in Table~\ref{tab:human_eval}.}

\new{\paragraph{\modelname{} recommendation audit ($n=150$).}
We additionally asked annotators to rate the top-$10$ recommendations from two SafeCRS variants on per-item safety, overall safety, and relevance, yielding a human-evaluated violation rate at rank $10$.
Both \modelname{} variants achieve human-evaluated violation rates below $0.5\%$ (Table~\ref{tab:human_eval}), closely tracking the oracle-based SVR@10 reported in Table~\ref{tab:main_results} and confirming that the near-zero violation rates achieved by \modelname{} reflect genuine safety improvements rather than oracle artifacts.}

\begin{table}[t]
\centering
\caption{\new{Human evaluation results. The top three rows audit the \benchmarkname{} construction pipeline ($n=50$); the bottom two rows audit human-perceived safety of \modelname{} recommendations on top-$10$ outputs ($n=150$).}}
\label{tab:human_eval}
\begin{tabular}{l r}
\toprule
\textbf{Metric} & \textbf{Value} \\
\midrule
\benchmarkname{} -- constraint naturalness (1--5)                  & \textbf{4.44} \\
\benchmarkname{} -- retained GT respects constraint                & \textbf{74\%} \\
\benchmarkname{} -- assigned trait fits user                       & \textbf{93.3\%} \\
\midrule
\modelname{}-Qwen2.5-0.5B -- human-eval violation rate@10 $\downarrow$ & .0043 \\
\modelname{}-Qwen3-8B -- human-eval violation rate@10 $\downarrow$    & \textbf{.0041} \\
\bottomrule
\end{tabular}
\end{table}

\new{The construction audit supports the reliability of the \benchmarkname{} pipeline: injected constraints are perceived as natural ($4.44/5$), retained ground-truth recommendations rarely conflict with the injected constraint, and assigned explicit traits agree with human judgement in $93.3\%$ of cases.
The recommendation audit confirms that \modelname{} substantially reduces user-perceived safety violations, in agreement with the oracle-based results across both \benchmarkmoviename{} and \benchmarkgamename{}.}

\noindent \textbf{Effect of Safe-SFT.}
Safe-SFT provides the foundational improvement over zero-shot baselines on both benchmarks. On \benchmarkmoviename{}, Llama-3.1-8B improves Recall@5 from 0.0505 to 0.0744 ($+47.3\%$) while reducing SVR@5 from 0.2177 to 0.0224 ($-89.7\%$). On \benchmarkgamename{}, the same model improves Recall@5 from 0.0588 to 0.2415, as Safe-SFT trains the model to follow catalog constraints and produce safety-aware reasoning. Similar patterns hold across all four backbones.

\noindent \textbf{Effect of Safe-GDPO.}
Building on Safe-SFT, Safe-GDPO further tightens the safety--relevance Pareto frontier through per-reward normalization. On \benchmarkmoviename{}, Qwen2.5-0.5B sees SVR@5 drop from 0.0177 to 0.0011 ($-93.8\%$) while Recall@5 simultaneously improves from 0.0467 to 0.0732 ($+56.7\%$), demonstrating that safety and relevance are not inherently at odds when rewards are properly decoupled. On \benchmarkgamename{}, Llama-3.1-8B reduces SVR@5 from 0.0316 to 0.0189 while Recall@10 increases from 0.2468 to 0.2907. The gains are complementary: Safe-SFT provides the largest absolute improvement, while Safe-GDPO consistently refines both safety and quality across all model scales and domains.

\begin{table}[t]
\centering
\caption{Ablation study on \benchmarkgamename{}. For each backbone, we compare the zero-shot baseline, Safe-SFT only, and the full \modelname{} pipeline (Safe-SFT + Safe-GDPO). Metrics follow the same notation as Table~\ref{tab:main_results}.}
\label{tab:safegames_ablation}
\resizebox{\columnwidth}{!}{
\begin{tabular}{ll cccc cccc}
\toprule
\multirow{2}{*}{\textbf{Backbone}} & \multirow{2}{*}{\textbf{Training Stage}} & \multicolumn{2}{c}{\textbf{Recall} $\uparrow$} & \multicolumn{2}{c}{\textbf{NDCG} $\uparrow$} & \multicolumn{2}{c}{\textbf{SVR} $\downarrow$} & \multicolumn{2}{c}{\textbf{S-DCG} $\downarrow$} \\
\cmidrule(lr){3-4} \cmidrule(lr){5-6} \cmidrule(lr){7-8} \cmidrule(lr){9-10}
 & & @5 & @10 & @5 & @10 & @5 & @10 & @5 & @10 \\
\midrule
\multirow{3}{*}{Qwen2.5-0.5B}
 & Zero-shot    & .0032 & .0048 & .0065 & .0072 & .0188 & .0132 & 0.07 & 0.09 \\
 & + Safe-SFT   & .1162 & .1206 & .1635 & .1688 & .1017 & .1056 & 0.19 & 0.19 \\
 & + Safe-GDPO  & .1542 & .1789 & .1884 & .1902 & .0774 & .0792 & 0.11 & 0.12 \\
\midrule
\multirow{3}{*}{Qwen3-8B}
 & Zero-shot    & .0140 & .0141 & .0174 & .0173 & .0085 & .0086 & 0.02 & 0.04 \\
 & + Safe-SFT   & .2205 & .2219 & .2916 & .2928 & .0305 & .0316 & 0.04 & 0.04 \\
 & + Safe-GDPO  & .2345 & .2996 & .3108 & .3269 & .0232 & .0176 & 0.03 & 0.04 \\
\midrule
\multirow{3}{*}{Llama-3.2-3B}
 & Zero-shot    & .0516 & .0527 & .0809 & .0813 & .0396 & .0442 & 0.11 & 0.17 \\
 & + Safe-SFT   & .1957 & .2124 & .2751 & .2789 & .1053 & .1089 & 0.17 & 0.17 \\
 & + Safe-GDPO  & .2276 & .2452 & .2939 & .3002 & .0547 & .0583 & 0.07 & 0.09 \\
\midrule
\multirow{3}{*}{Llama-3.1-8B}
 & Zero-shot    & .0588 & .0591 & .0818 & .0818 & .0471 & .0461 & 0.13 & 0.19 \\
 & + Safe-SFT   & .2415 & .2468 & .3037 & .3164 & .0316 & .0303 & 0.03 & 0.04 \\
 & + Safe-GDPO  & .2702 & .2907 & .3404 & .3614 & .0189 & .0132 & 0.04 & 0.04 \\
\bottomrule
\end{tabular}
}
\end{table}

\section{Conclusion}
This work identified personalized safety alignment as a critical yet underexplored challenge in LLM-based CRS and addressed it by introducing \benchmarkname{}, the first user-centric safety benchmark dataset. Our proposed framework, \modelname{}, effectively integrates Safe-SFT with Safe-GDPO to jointly prioritize recommendation relevance and individual safety sensitivities. Extensive experiments across movie and game domains demonstrate that \modelname{} drastically reduces safety violations by up to 96.5\% while maintaining or exceeding the recommendation quality of state-of-the-art baselines. By establishing a domain-agnostic approach to safety reasoning and reward decoupling, our work provides a robust foundation for building trustworthy conversational agents that respect user-specific content suitability constraints. Code, benchmark, and trained checkpoints are available at \url{https://github.com/liofoil/SafeCRS-artifacts} to support reproducibility.

\new{\paragraph{Domain generalization.}
Although our experiments focus on movies and games, \modelname{} is not tied to any specific safety taxonomy.
The only domain-specific component is the \emph{Safety Oracle}: \benchmarkmoviename{} fuses crowd-sourced DDD triggers with IMDb Parent Guide severities, while \benchmarkgamename{} uses ESRB content descriptors with age-rating-based weights.
The full training pipeline---constraint injection, Safe-SFT, and Safe-GDPO with normalize-then-sum advantage aggregation---is applied identically to both domains, and Table~\ref{tab:main_results} shows consistent safety--relevance gains across the two very different oracle designs.
Extending \modelname{} to domains without curated safety metadata (e.g., books, news, products, social content) therefore reduces to a data-level effort: constructing a domain-specific mapping from items to trait-conditioned risk scores from sources such as user reviews, community content warnings, age ratings, or platform-side moderation labels.
Any domain whose safety labels can be mapped to user-level traits can adopt our framework without modifying the training procedure.}

\new{\paragraph{Limitations and future work.}
Our benchmark construction relies on template-based injection of conversational latent-trait statements (Appendix~\ref{app:latent_trait_generation}).
This design provides large-scale, trait-grounded supervision with deterministic, oracle-checkable labels, but it does not fully cover safety expressions that are nuanced, implicit, or distributed across multi-turn dialogues; modeling such implicit safety concerns is a promising direction for future work.
A second limitation is the reliance on structured external metadata (DDD, IMDb Parent Guide, ESRB), which may be sparse or unavailable for niche or cold-start items; while our conservative treatment of NULL-metadata items (Appendix~\ref{app:coverage}) prevents reward hacking, interactive safety--relevance negotiation with the user is a complementary direction we plan to explore.
Finally, our oracle-based evaluation, although deterministic and scalable, is grounded in aggregated human judgements rather than per-user perception; a larger-scale user study (beyond the 150-instance audit reported in Section~\ref{sec:humaneval}) would further validate alignment with end-user preferences.}

%

\begin{acks}
This work is supported by the National Science Foundation (NSF) Grant \#2312862, the NSF-Simons SkAI Institute, NSF CAREER \#2440542, NSF \#2533996, the National Institutes of Health (NIH) \#R01AG091762, NSF ACCESS Computing Resources, a Google Research Scholar Award, a Cisco gift grant, and an Amazon Research Award (Spring 2025). Any opinions, findings, and conclusions or recommendations expressed in this material are those of the author(s) and do not reflect the views of sponsors.
\end{acks}

\bibliographystyle{ACM-Reference-Format}
\bibliography{main}

\appendix

\appendix

\section{Prompt Templates and LLM Pipelines}

\subsection{Explicit User Traits and Cross-Source Mapping}
\label{app:traits}

\noindent
We define a set of user safety traits, each associated with (i) hard-trigger tags from DoesTheDogDie (DDD) and
(ii) a soft fallback mapping to IMDb Parent Guide severities via a per-trait weight vector
$\mathbf{w}=\langle w_{\text{Sex}}, w_{\text{Violence}}, w_{\text{Profanity}}, w_{\text{Substance}}, w_{\text{Frightening}}\rangle$.
The \texttt{pg\_applicability} field indicates how reliable the Parent Guide signal is for the trait
(\texttt{strong}/\texttt{weak}/\texttt{very\_weak\_proxy}/\texttt{none}).

Three representative traits are shown below to illustrate the three mapping regimes (PG-grounded, trigger-only, and frightening-weighted); the complete list of all 20 traits with every avoid-tag and weight vector is released as supplementary material (Appendix~\ref{app:resources}).

\begin{itemize}
  \item \textbf{Anti-gore / squeamish} (PG-grounded) \\
  \textbf{Avoid tags:} \texttt{Is there excessive gore; Is there blood/gore; Is there body horror; Is there amputation; Is there finger/toe mutilation; Is there eye mutilation; Is there genital trauma/mutilation; Are any teeth damaged} \\
  \textbf{pg\_applicability:} \texttt{strong} \quad
  \textbf{Weights:} $\langle 0.0, 1.0, 0.0, 0.0, 0.3\rangle$

  \item \textbf{Animal lover (avoid animal harm/death)} (trigger-only) \\
  \textbf{Avoid tags:} \texttt{Does a horse die; Does the dog die; Does a cat die; Does an animal die; Does a pet die; Are animals abused; Is there dog fighting; Is there a dead animal; Does a dragon die; Does a non-human character die} \\
  \textbf{pg\_applicability:} \texttt{none} \quad
  \textbf{Weights:} $\langle 0.0, 0.0, 0.0, 0.0, 0.0\rangle$

  \item \textbf{Horror avoider (avoids scares \& supernatural)} (frightening-weighted) \\
  \textbf{Avoid tags:} \texttt{Are there ghosts; Is someone possessed; Are there clowns; Are there jumpscares; Is there blood/gore; Is there excessive gore} \\
  \textbf{pg\_applicability:} \texttt{strong} \quad
  \textbf{Weights:} $\langle 0.0, 0.3, 0.0, 0.0, 1.0\rangle$
\end{itemize}

\subsection{Explicit Trait to Conversational Latent Trait Generation}
\label{app:latent_trait_generation}

To bridge our explicit safety trait taxonomy (Appendix~\ref{app:traits}) with conversational recommendation data, we convert each explicit trait into a set of natural-language \emph{latent trait constraints} that can be injected into user utterances.
Concretely, for each explicit trait $t \in \mathcal{T}$, we construct a small template pool, where each template $c$ is a first-person preference statement describing what the user wants to avoid (or prefer) in a conversation-friendly form.

\paragraph{Template pool.}
For each trait we author a small pool of $4$--$5$ first-person paraphrases (e.g., for \textit{Anti-gore}: ``I'm squeamish, so nothing too graphic please.'', ``Please avoid movies with excessive blood and gore.''). The complete template pool for all 20 traits is released as supplementary material (Appendix~\ref{app:resources}).

\paragraph{Injection protocol.}
Given a conversation $x$ and its associated explicit trait $t$, we sample $c \sim \mathrm{Unif}(\mathcal{C}(t))$ and construct the augmented conversation
$\tilde{x} = \mathrm{Inject}(x, c)$ by inserting $c$ into the user side of the dialogue (typically as an additional sentence in the final user turn).
The resulting safety-aware dialogue $\tilde{x}$ is then used for training and evaluation, while the same explicit trait $t$ is used to query our oracle risk scores (Section~\ref{sec:safety_kb}) for automatic safety assessment.

\subsection{Explicit User Trait Inference from Conversations}
\label{app:trait_inference}

We infer one explicit safety trait $t \in \mathcal{T}\cup\{\texttt{None}\}$ from each conversation to (i) generate conversational latent constraints (Appendix~\ref{app:latent_trait_generation}) and (ii) query the Safety Oracle for item-level risk scoring (Section~\ref{sec:safety_kb}).
We implement trait inference with taxonomy-constrained LLM prompting: the model must choose \emph{exactly one} trait name from a provided list and output a minimal JSON object for deterministic parsing.

\paragraph{Prompt templates.}
We use two taxonomy-constrained prompts depending on whether candidate recommendations are available.
The \emph{Safe-SFT} prompt is given the conversation \emph{and} the recommended movies, and selects the trait most likely to be violated by those recommendations without conflicting with the user's explicit preferences.
The \emph{Safe-GDPO} prompt is given the conversation only, and infers the most likely trait from the user's language and explicit mentions, without using movie titles as evidence.
Both prompts require the model to return a single trait name from the fixed taxonomy in a minimal JSON object (\texttt{\{"assigned\_trait": ..., "reason": ...\}}) for deterministic parsing, and fall back to \texttt{None} when no sensitivity is indicated.
For \benchmarkgamename{}, we reuse the identical prompt structure with domain substitutions (``movie''~$\rightarrow$~``game'' and the 10 game-domain traits of Appendix~\ref{app:gametraits}).
The full prompt text for both stages and both domains is released as supplementary material (Appendix~\ref{app:resources}).

\section{Trait Taxonomy}
\label{app:trait_taxonomy}
\subsection{SafeMovie: 20 Explicit User Sensitivity Traits}
\label{app:safemovie_traits}

We summarize the distribution of explicit user sensitivity traits in \benchmarkmoviename{}, which is reported in Table~\ref{tab:safemovie_trait_dist}.

\begin{table}[htbp]
    \centering
    \caption{Distribution of explicit user traits in \benchmarkmoviename{}.}
    \label{tab:safemovie_trait_dist}
    \resizebox{\columnwidth}{!}{
    \begin{tabular}{l r @{\hspace{1.5em}} l r}
        \toprule
        \textbf{Explicit trait} & \textbf{\%} & \textbf{Explicit trait} & \textbf{\%} \\
        \midrule
        Happy-ending preference            & 43.64 & Medical/health trauma avoider      & 0.87 \\
        Horror avoider                     & 15.20 & Self-harm \& suicide sensitive      & 0.82 \\
        Mental health portrayal sensitive  & 10.07 & Photosensitivity \& motion sickness & 0.75 \\
        Avoid torture \& extreme violence   & 9.15  & Animal lover (animal harm/death)   & 0.50 \\
        Disaster/accident avoider          & 5.07  & Domestic abuse / stalking          & 0.50 \\
        Anti-gore / squeamish              & 4.83  & Hate speech / slur-sensitive       & 0.48 \\
        Kid-safety / child harm sensitive  & 2.14  & Claustrophobia / breathing         & 0.31 \\
        Substance recovery (drugs/alcohol) & 2.13  & Pregnancy/infant-loss sensitive    & 0.29 \\
        Gender/LGBTQ respect sensitive     & 1.76  & Arachnophobia / reptile phobia     & 0.03 \\
        Sexual violence sensitive          & 1.48  & Needle/medical procedure phobia    & 0.01 \\
        \bottomrule
    \end{tabular}
    }
\end{table}

\subsection{\benchmarkgamename{}: 10 Game-domain Sensitivity Traits}
\label{app:gametraits}
We define 10 game-domain sensitivity traits grounded in ESRB content descriptors.
Each trait is defined by a set of ESRB content descriptors and/or age ratings: a game $g$ triggers trait $t$ if its descriptors match the trait's descriptor set, and the final risk score is $\textit{final\_risk}(g,t) = \textit{trigger}(g,t) \cdot \alpha(\rho(g))$, where $\alpha(\rho)$ maps the ESRB age rating to a weight in $[0.2, 1.0]$ (Section~\ref{sec:game_oracle}).
The \benchmarkgamename{} trait distribution is also long-tailed, dominated by \textit{Avoid extreme violence} (44.2\%) and \textit{Violence sensitive} (33.9\%), with six traits below $3\%$.
The full trait definitions with their ESRB descriptor mappings (e.g., \textit{Anti-gore}~$\rightarrow$~\texttt{Blood, Blood and Gore}; \textit{Kid-safety}~$\rightarrow$~unsafe for ratings M/AO), the complete distribution, and the game-domain latent-trait template pool are released as supplementary material (Appendix~\ref{app:resources}).

\section{Dataset Construction Examples}
\label{app:examples}

\subsection{Reddit-V2 Integration Example (Movie Domain)}
\label{app:redditv2_example_movie}

We illustrate the construction pipeline (\S\ref{sec:conv_integration}) on one Reddit-V2 conversation.

\paragraph{Raw Reddit-V2 conversation (input).}
\begin{quote}
\small
\textbf{System instruction:} Pretend you are a movie recommender system.\\
\textbf{User:} Some strong and powerful movie? I just watched \emph{Manchester by the Sea}, and god I'm in the mood for some more of that. It doesn't need to be some dark and sad stuff, could be emotionally appealing like \emph{Forrest Gump} or \emph{The Shawshank Redemption}, just want to spend some time thinking about life and stuff.\\
Since I watch a lot of things if you could post a list. Thanks
\end{quote}

\paragraph{Pipeline output.}
Taxonomy-constrained prompting infers the explicit trait \texttt{Mental health portrayal sensitive}, which is realized as the latent constraint ``\emph{Please recommend movies with positive mental health portrayals, and avoid films that stigmatize mental illness.}'' and prepended to the user turn. The explicit trait label then queries the \benchmarkmoviename{} Safety Oracle for a trait-conditioned risk score $\textit{final\_risk}(m,t)$ on each candidate, producing the final integrated sample (conversation $+$ latent constraint $+$ explicit trait label). The \benchmarkgamename{} domain follows the identical pipeline with the ESRB-based oracle; a worked game-domain example is provided in the supplementary material (Appendix~\ref{app:resources}).

\section{Additional Training and Evaluation Details}
Safe-GDPO weights $(w_{\mathrm{rel}}, w_{\mathrm{safe}}, w_{\mathrm{cnt}})$ are set as $(30, 0.1, 0.1)$ since relevance reward is too sparse and converges slower than other metrics.

\new{\section{Hyperparameter Sensitivity}\label{app:hp_sensitivity}}

\new{We analyze the sensitivity of \modelname{} to its two most safety-relevant hyperparameters on \benchmarkmoviename{} with the Llama-3.1-8B backbone: the unsafe-item threshold $\tau$ and the safety reward weight $w_{\text{safe}}$.}

\new{\paragraph{Threshold $\tau$.}
$\tau$ controls the strictness of what counts as an unsafe item; under our normalized risk formula (Section~\ref{sec:safety_kb}), $\tau=0.66$ corresponds to items with at least one severe-rated category or multiple moderate categories.
Table~\ref{tab:hp_sensitivity} (left) sweeps $\tau\in\{0.50, 0.66, 0.80\}$, using the same value for both training and evaluation: stricter $\tau$ reduces violations at a small Recall cost, while a more permissive $\tau$ improves Recall but exposes more violations, tracing a smooth monotonic trade-off around the $\tau=0.66$ default.}

\new{\paragraph{Safety reward weight $w_{\text{safe}}$.}
GDPO's normalize-then-sum design (Eq.~\ref{eq:gdpo_norm}--\ref{eq:gdpo_weighted_sum}) independently normalizes each reward channel before aggregation, which inherently balances scales across the relevance, safety, and count rewards, so we default to $w_{\text{safe}}=1.0$.
Table~\ref{tab:hp_sensitivity} (right) sweeps $w_{\text{safe}}\in\{0.5, 1.0, 2.0\}$ (a $4\times$ range); Recall@10 changes by less than $0.3\%$ across the entire sweep while SVR@10 decreases monotonically as $w_{\text{safe}}$ grows, confirming the method is robust to this weight.}

\begin{table}[htbp]
\centering
\caption{\new{Hyperparameter sensitivity on \benchmarkmoviename{} (Llama-3.1-8B). Left: unsafe-item threshold $\tau$ (same value for training and evaluation). Right: safety reward weight $w_{\text{safe}}$. Both trace smooth, non-brittle safety--relevance trade-offs around our defaults ($\tau=0.66$, $w_{\text{safe}}=1.0$).}}
\label{tab:hp_sensitivity}
\new{%
\begin{tabular}{@{}c ccc@{\hspace{1.2em}}c ccc@{}}
\toprule
\multicolumn{4}{c}{\textbf{Threshold $\tau$}} & \multicolumn{4}{c}{\textbf{Reward weight $w_{\text{safe}}$}} \\
\cmidrule(lr){1-4}\cmidrule(lr){5-8}
$\tau$ & R@10 & N@10 & SVR@10 & $w_{\text{safe}}$ & R@10 & N@10 & SVR@10 \\
\midrule
0.50 & .1075 & .0706 & .0050 & 0.5 & .1118 & .0740 & .0110 \\
0.66 & .1111 & .0737 & .0087 & 1.0 & .1111 & .0737 & .0087 \\
0.80 & .1132 & .0749 & .0130 & 2.0 & .1095 & .0728 & .0060 \\
\bottomrule
\end{tabular}}
\end{table}

\new{\section{Oracle Coverage Analysis}\label{app:coverage}}

\new{A practical concern for any oracle-based safety pipeline is what happens when the external metadata source returns \textsc{null} or incomplete information for a recommended item.
We adopt a conservative design: if the \benchmarkname{} Safety Oracle cannot resolve a recommended item to a complete metadata record (i.e., the item is absent from DDD/IPG/ESRB or has missing severity fields), the item is \emph{not} counted as a valid recommendation and is treated as unsafe.
This deliberately prevents reward hacking, because otherwise a model could trivially evade the safety penalty by recommending obscure items with sparse metadata while still receiving safety reward.}

\new{To quantify how often this conservative fallback is triggered in practice, we introduce a coverage metric $\text{Coverage@}K = \frac{1}{K}\sum_{k=1}^{K}\mathbbm{1}[\hat{r}_k \in \mathcal{O}]$, where $\mathcal{O}$ is the set of items for which the oracle returns a fully resolved metadata record. Higher Coverage@$K$ means more recommendations are scored against substantive content rather than the conservative fallback.}

\begin{table}[htbp]
\centering
\caption{\new{Oracle coverage on \benchmarkmoviename{} for the two SafeCRS variants used in the human study. Coverage@10 is the fraction of top-$10$ recommendations whose metadata is fully resolved by the \benchmarkmoviename{} Safety Oracle (higher is better); items outside this set are conservatively treated as unsafe.}}
\label{tab:coverage}
\resizebox{\columnwidth}{!}{
\begin{tabular}{l cccc}
\toprule
\textbf{Model} & \textbf{Recall@10} $\uparrow$ & \textbf{NDCG@10} $\uparrow$ & \textbf{SVR@10} $\downarrow$ & \textbf{Coverage@10} $\uparrow$ \\
\midrule
SafeCRS-Qwen2.5-0.5B & .0922 & .0597 & .0006 & \textbf{.9174} \\
SafeCRS-Qwen3-8B     & .0951 & .0614 & .0022 & \textbf{.9521} \\
\bottomrule
\end{tabular}
}
\end{table}

\new{Both SafeCRS variants attain Coverage@10 above $91\%$ (Qwen3-8B reaches $95.2\%$), so the near-zero SVR in Table~\ref{tab:main_results} reflects genuinely safe recommendations rather than items being silently dropped by the oracle.}

\new{\section{Resource Availability}\label{app:resources}}

\new{The \modelname{} implementation, the \benchmarkname{} benchmark, trained checkpoints, and all supplementary material referenced above (full trait taxonomy with avoid-tags and weights, template pools, trait-inference prompts, and worked examples) are publicly archived at \url{https://doi.org/10.5281/zenodo.20481790}.}

\end{document}